\documentclass[runningheads]{llncs}
\usepackage{amssymb}
\usepackage{graphicx}
\begin{document}
\title{Training behavior of deep neural network in frequency domain\thanks{Originally submitted to arxiv.org on 3 Jul 2018. To appear in  2019 26th-International conference of neural information processing (ICONIP).}}
%
%
\author{Zhi-Qin John Xu\thanks{Corresponding author}\inst{1} \and Yaoyu Zhang\inst{2} \and
Yanyang Xiao\inst{2}}
\authorrunning{Zhi-Qin J. Xu, Yaoyu Zhang, Yanyang Xiao}
\institute{School of Mathematical Sciences and Institute of Natural Sciences, Shanghai Jiao Tong University, Shanghai, China,\\
\email{xuzhiqin@sjtu.edu.cn},
\and
School of Mathematics, Institute for Advanced Study, Princeton, NJ 08540, USA,\\
\email{yaoyu@ias.edu},
\and The Brain Cognition and Brain Disease Institute, Shenzhen Institutes of Advanced Technology, Chiniese Academy of Sciences, China,\\ \email{xyy82148@gmail.com}}
\maketitle              
\begin{abstract}
Why deep neural networks (DNNs) capable of overfitting often generalize
well in practice is a mystery \cite{zhang2016understanding}. To find
a potential mechanism, we focus on the study of implicit biases underlying
the training process of DNNs. In this work, for both real and synthetic
datasets, we empirically find that a DNN with common settings first
quickly captures the dominant low-frequency components, and then relatively
slowly captures the high-frequency ones. We call this phenomenon Frequency
Principle (F-Principle). The F-Principle can be observed over DNNs
of various structures, activation functions, and training algorithms
in our experiments. We also illustrate how the F-Principle helps understand
the effect of early-stopping as well as the generalization of DNNs.
This F-Principle potentially provides insight into a general principle
underlying DNN optimization and generalization.

\end{abstract}

\section{Introduction}

Although Deep Neural Networks (DNNs) are totally transparent, i.e.,
the value of each node and each parameter can be easily obtained,
it is difficult to interpret how information is processed through
DNNs. We can easily record the trajectories of the parameters of DNNs
during the training. However, it remains unclear what is the general
principle underlying the highly non-convex problem of DNN optimization
\cite{lecun2015deep}. Therefore, DNN is often criticized for being
a ``black box'' \cite{alain2016understanding,shwartz2017opening}.
Even for the simple problem of fitting one-dimensional (1-d) functions,
the training process of DNN is still not well understood \cite{Saxe2018Information,wu2017towards}.
For example, Wu et al. (2017) \cite{wu2017towards} use DNNs of different
depth to fit a few data points sampled from a 1-d target function
of third-order polynomial. They find that, even when a DNN is capable
of over-fitting, i.e., the number of its parameters is much larger
than the size of the training dataset, it often generalizes well (i.e.,
no overfitting) after training. In practice, the same phenomenon is
also observed for much more complicated datasets \cite{kawaguchi2017generalization,martin2017rethinking,zheng2017understanding,zhang2016understanding,lin2016generalization,wu2017towards}.
Intuitively, for a wide DNN, its solutions of zero training error
lies in a huge space where well-generalized ones only occupy a small
subset. Therefore, it is mysterious that DNN optimization often ignores
a huge set of over-fitting solutions. To find an underlying mechanism,
in this work, we characterize the behavior of the DNN optimization
process in the frequency domain using 1-d functions as well as real
datasets of image classification problems (MNIST and CIFAR10). Our
work provides insight into an implicitly bias underlying the training
process of DNNs. 

We empirically find that, for real datasets or synthetic functions,
a DNN with common settings first quickly captures their dominant low-frequency
components while keeping its own high-frequency ones small, and then
relatively slowly capture their high-frequency components. We call
this phenomenon \emph{Frequency Principle } (F-Principle). From our numerical experiments, this F-Principle
can be widely observed for DNNs of different width (tens to thousands
of neurons in each layer), depth (one to tens of hidden layers), training
algorithms (gradient descent, stochastic gradient descent, Adam) and
activation functions (tanh and ReLU). Remark that this strategy of
the F-Principle, i.e., fitting the target function progressively in
ascending frequency order, is also adopted explicitly in some numerical
algorithms to achieve remarkable efficiency. These numerical algorithms
include, for example, the Multigrid method for solving large-scale
partial differential equations \cite{hackbusch2013multi} and a recent
numerical scheme that efficiently fits the three-dimensional structure
of proteins and protein complexes from noisy two-dimensional images
\cite{barnett2017rapid}. 

The F-Principle provides a potential mechanism of why DNNs often generalize
well empirically albeit its ability of over-fitting \cite{zhang2016understanding}.
For a finite training set, there exists an effective frequency range
\cite{shannon1949communication,yen1956nonuniform,mishali2009blind}
beyond which the information of the signal is lost. By the F-Principle,
with no constraint on the high-frequency components beyond the effective
frequency range, DNNs tend to keep  them small. For a wide class of
low-frequency dominant natural signals (e.g., image and sound), this
tendency coincides with their behavior of decaying power at high frequencies.
Thus, DNNs often generalize well in practice. When the training data
is noisy, the small-amplitude high-frequency components are easier
to be contaminated. By the F-Principle, DNNs first capture the less
noisy low-frequency components of the training data and keep higher-frequency
components small. At this stage, although the loss function is not
best optimized for the training data, DNNs could generalize better
for not fitting the noise dominating the higher-frequencies. Therefore,
as widely observed, early-stopping often helps generalization.

Our key contribution in this work is the discovery of an F-Principle
underlying the training of DNNs for both synthetic and real datasets.
In addition, we demonstrate how the F-Principle provides insight into
the effectiveness of early stopping and the good generalization of
DNNs in general.

\section{Related works}

Consistent with other studies \cite{wu2017towards,arpit2017closer},
our analysis shows that over-parameterized DNNs tend to fit training
data with low-frequency functions, which are naturally of lower complexity.
Intuitively, lower-frequency functions also possess smaller Lipschitz
constants. According to the study in Hardt et al. (2015) \cite{hardt2015train},
which focuses on the relation between stability and generalization,
smaller Lipschitz constants can lead to smaller generalization error. 

The F-Principle proposed in this work initiates a series of works
\cite{xu2018understanding,xu2018frequency,xu2019frequency,rabinowitz2019meta,zhen2018nonlinear,zhang_LFP_2019,cai2019phasednn}.
A stronger verification of the F-Principle for the high dimensional
datasets can be found in Xu et al., (2019) \cite{xu2019frequency}.
Theoretical studies on the F-Principle can be found in Xu et al.,
(2019), Xu et al., (2019) and Zhang et al., (2019) \cite{xu2018understanding,xu2019frequency,zhang_LFP_2019}.
The F-Principle is also used as an important phenomenon to pursue
fundamentally different learning trajectories of meta-learning \cite{rabinowitz2019meta}.
The theoretical framework \cite{xu2018understanding,xu2019frequency}
of analyzing the F-Principle is used to analyze a nonlinear collaborative
scheme for deep network training \cite{zhen2018nonlinear}. Based
on the F-Principle, a fast algorithm by shifting high frequencies
to lower ones is developed for fitting high frequency functions \cite{cai2019phasednn}.
These subsequent works show the importance of the F-Principle.

\section{Experimental setup}

We summarize the setups for each figure as follows. All DNNs are trained
by the Adam optimizer, whose parameters are set to their default values
\cite{kingma2014adam}. The loss function is the mean-squared error.
The parameters of DNNs are initialized by a Gaussian distribution
with mean $0$.

In Fig. \ref{fig:CNNDNN_MC}, the setting of the fully-connected tanh-DNN
is as follows. Width of hidden layers: 200-100-100;  Batch size: 100;
Learning rate: $10^{-5}$ for CIFAR10 and $10^{-6}$ for MNIST; Standard
deviation of Gaussian initialization: $10^{-4}$. The setting of the
ReLU-CNN is as follows: two layers of $32$ features with $3\times3$
convolutional kernel and $2\times2$ max-pooling, followed by 128-64
densely connected layers; Batch size: 128; Standard deviation of Gaussian
initialization: $0.05$. We select $10000$ samples from each dataset
for the training.

In Figs. (\ref{fig:sinx}, \ref{fig:Generalization}, \ref{fig:Init},
\ref{fig:ERJnum}), we use a fully-connected tanh-DNN of $4$ hidden
layers of width 200-200-200-100, standard deviation of Gaussian initialization
$0.1$, learning rate $2\times10^{-5}$ and full-batch size training.

In addition, $\mathcal{F}[\cdot]$ indicates the Fourier transform,
which is experimentally estimated on discrete training or test data
points.

\section{F-Principle }

In this section, we study the training process of DNNs in the frequency
domain. We empirically find that, for a general class of functions
dominated by low-frequencies, the training process of DNNs follows
the F-Principle by which low-frequency components are first captured,
followed by high-frequency ones. 

\subsection{MNIST/CIFAR10 \label{sec:Verification-in-natural-1}}

Since the computation of high-dimensional Fourier transform suffers
from the curse of dimensionality, to verify the F-Principle in the
image classification problems (MNIST and CIFAR10), we perform the
Fourier analysis along the first principle component of the training
inputs. 

The training set is a list of labeled images denoted by $\{\vec{x}_{k};y_{k}\}_{k=0}^{n-1}$,
where each image $\vec{x}_{k}\in[0,1]^{N_{in}}$, $N_{in}$ is the
number of pixels of an image, each label $y_{k}\in\{0,1,2,\cdots9\}$.
We use DNNs of two structures to learn this training set, that is,
a fully-connected DNN and a CNN. Denote $x_{k}=\vec{x}_{k}\cdot\vec{v}_{PC}$,
which is the projection of image $\vec{x}_{k}$ along the direction
of the first principle component of $\{\vec{x}_{k}\}_{k=0}^{n-1}$
denoted by a unit vector $\vec{v}_{PC}$. Using non-uniform Fourier
transform, we obtain
\[
\mathcal{F}_{PC}^{n}[y](\gamma)=\frac{1}{n}\sum_{j=0}^{n-1}y_{j}\exp\left(-2\pi ix_{j}\gamma\right),
\]
where $\gamma\in\mathbb{Z}$ is the \emph{frequency index}. For the
DNN output $T(\vec{x}_{k})$, similarly, $\mathcal{F}_{PC}^{n}[T](\gamma)=\frac{1}{n}\sum_{j=0}^{n-1}T(\vec{x}_{j})\exp\left(-2\pi ix_{j}\gamma\right)$.
To examine the convergence behavior of different frequency components
during the training of a DNN, we compute the relative difference of
$\mathcal{F}_{PC}^{n}[T][\gamma]$ and $\mathcal{F}_{PC}^{n}[y][\gamma]$
at each recording step, i.e., 
\begin{equation}
\Delta_{F}(\gamma)=\frac{|\mathcal{F}_{PC}^{n}[y](\gamma)-\mathcal{F}_{PC}^{n}[T](\gamma)|}{|\mathcal{F}_{PC}^{n}[y](\gamma)|},\label{eq:dfreq}
\end{equation}
 where $|\cdot|$ denotes the absolute value. As shown in the first
column in Fig. \ref{fig:CNNDNN_MC}, both datasets are dominated by
low-frequency components along the first principle direction. Theoretically,
frequency components other than the peaks are susceptible to the artificial
periodic boundary condition implicitly applied in the Fourier transform,
thereby are not essential to our frequency domain analysis \cite{percival1993spectral}.
In the following, we only focus on the convergence behavior of the
frequency peaks during the training. By examining the relative error
of certain selected key frequency components (marked by black squares),
one can clearly observe that DNNs of both structures for both datasets
tend to capture the training data in an order from low to high frequencies
as stated by the F-Principle \footnote{Almost at the same time, another research \cite{rahaman2018spectral}
finds a similar result. However, they add noise to MNIST,
which contaminates the labels. } (second and third column in Fig. \ref{fig:CNNDNN_MC}).
\begin{center}
\begin{figure}[h]
\begin{centering}
\includegraphics[scale=0.45]{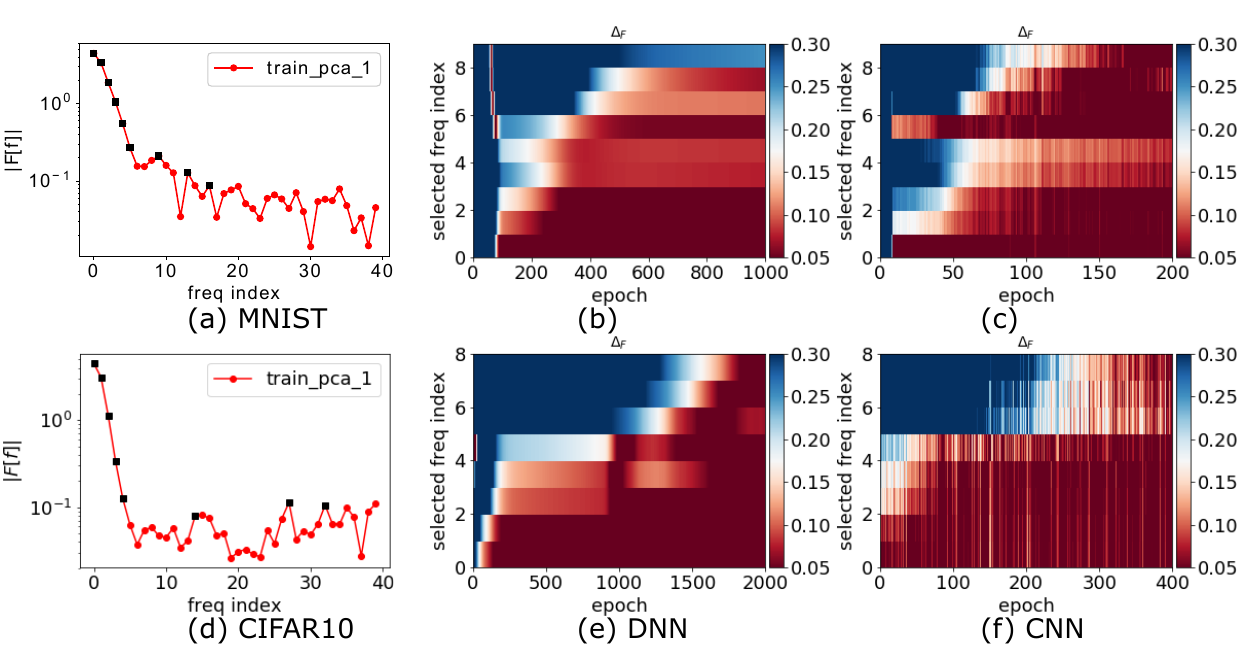}
\par\end{centering}
\caption{Frequency analysis of DNN output function along the first principle
component during the training. The training datasets for the first
and the second row are from MNIST and CIFAR10, respectively. The neural
networks for the second column and the third column are fully-connected
DNN and CNN, respectively. (a,d) $|\mathcal{F}_{PC}^{n}[y](\gamma)|$.
The selected frequencies are marked by black dots. (b, c, e, f) $\Delta_{F}$
at different recording epochs for different selected frequencies.
$\Delta_{F}$ larger than $0.3$ (or smaller than $0.05$) is represented
by blue (or red). \label{fig:CNNDNN_MC}}
 
\end{figure}
\par\end{center}

\subsection{Synthetic data}

In this section, we demonstrate the F-Principle by using synthetic
data sampled from a target function of known intrinsic frequencies.
We design a target function by discretizing a smooth function $f_{0}(x)$
as follows,
\begin{equation}
y=f(x)=\alpha\times{\rm Round}(f_{0}(x)/\alpha),\quad\alpha\in(0,\infty),\label{eq:fittingFunc}
\end{equation}
where ${\rm Round(\cdot)}$ takes the nearest integer value. We define
$y=f_{0}(x)$ for $\alpha=0$. We consider $f_{0}(x)=\sin(x)+2\sin(3x)+3\sin(5x)$
with $\alpha=2$ as shown in Fig. \ref{fig:sinx}a. As shown in Fig.
\ref{fig:sinx}b, for the discrete Fourier transform (DFT) of $f(x)$,
i.e., ${\cal F}[f]$, there are three most important frequency components
and some small peaks due to the discretization. In this case, we can
observe a precise convergence order from low- to high-frequency for
frequency peaks as shown in Fig. \ref{fig:sinx}c. 

We have performed the same frequency domain analysis for various low-frequency
dominant functions, such as $f_{0}(x)=|x|$, $f_{0}(x)=x^{2}$ and
$f_{0}(x)=\sin(x)$ with different $\alpha$'s (results are not shown),
for both ReLU and tanh activation functions, and both gradient descent
and Adam \cite{kingma2014adam} optimizers. We find that F-Principle
always holds during the training of DNNs. Therefore, the F-Principle
seems to be an intrinsic character of DNN optimization. 
\begin{center}
\begin{figure*}
\begin{centering}
\includegraphics[scale=0.4]{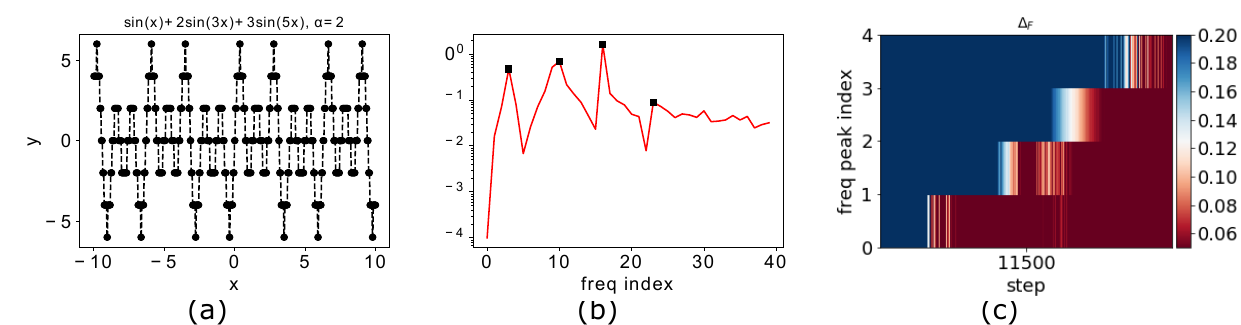}
\par\end{centering}
\caption{Frequency domain analysis of the training process of a DNN for $f_{0}(x)=\sin(x)+2\sin(3x)+3\sin(5x)$
with $\alpha=2$ in Eq. (\ref{eq:fittingFunc}). (a) The target function.
(b) $|\mathcal{F}[f]|$ at different frequency indexes. First four
frequency peaks are marked by black dots. (c) $\Delta_{F}$ at different
recording steps for different frequency peaks. The training data is
evenly sampled in $[-10,10]$ with sample size $600$. \label{fig:sinx}}
 
\end{figure*}
\par\end{center}

\section{Understanding the training behavior of DNNs by the F-Principle \label{sec:F-Principle-in-understanding}}

In this section, we provide an explanation based on the F-Principle
of why DNNs capable of over-fitting often generalize well in practice
\cite{kawaguchi2017generalization,martin2017rethinking,zheng2017understanding,zhang2016understanding,lin2016generalization,wu2017towards}.
For a class of functions dominated by low frequencies, with finite
training data points, there is an \emph{effective frequency range}
for this training set, which is defined as the range in frequency
domain bounded by Nyquist-Shannon sampling theorem \cite{shannon1949communication}
when the sampling is evenly spaced, or its extensions \cite{yen1956nonuniform,mishali2009blind}
otherwise. When the number of parameters of a DNN is greater than
the size of the training set, the DNN can overfit these sampling data
points (i.e., training set) with different amount of powers outside
the effective frequency range. However, by the F-Principle, the training
process will implicitly bias the DNN towards a solution with a low
power at the high-frequencies outside the effective frequency range.
For functions dominated by low frequencies, this bias coincides with
their intrinsic feature of low power at high frequencies, thus naturally
leading to a well-generalized solution after training. By the above
analysis, we can predict that, in the case of insufficient training
data, when the higher-frequency components are not negligible, e.g.,
there exists a significant frequency peak above the effective frequency
range, the DNN cannot generalize well after training. 

In another case where the training data is contaminated by noise,
early-stopping method is usually applied to avoid overfitting in practice
\cite{lin2016generalization}. By the F-Principle, early-stopping
can help avoid fitting the noisy high-frequency components. Thus,
it naturally leads to a well-generalized solution. We use the following
example for illustration.

As shown in Fig. \ref{fig:Generalization}a, we consider $f_{0}(x)=\sin(x)$
with $\alpha=0.5$ in Eq. (\ref{eq:fittingFunc}). For each sample
$x$, we add a noise $\epsilon$ on $f_{0}(x)$, where $\epsilon$
follows a Gaussian distribution with mean $0$ and standard deviation
$0.1$. The DNN can well fit the sampled training set as the loss
function of the training set decreases to a very small value (green
stars in Fig. \ref{fig:Generalization}b). However, the loss function
of the test set first decreases and then increases (red dots in Fig.
\ref{fig:Generalization}b). That is, the generalization performance
of the DNN gets worse during the training after a certain step. In
Fig. \ref{fig:Generalization}c, $|\mathcal{F}[f]|$ for the training
data (red) and the test data (black) only overlap around the dominant
low-frequency components. Clearly, the high-frequency components of
the training set are severely contaminated by noise. Around the turning
step --- where the best generalization performance is achieved, indicated
by the green dashed line in Fig.  \ref{fig:Generalization}b ---
the DNN well captures the dominant peak as shown in Fig.  \ref{fig:Generalization}c.
After that, clearly, the loss function of the test set increases as
DNN start to capture the higher-frequency noise (red dots in Fig.
\ref{fig:Generalization}b). These phenomena conform with our analysis
that early-stopping can lead to a better generalization performance
of DNNs as it helps prevent fitting the noisy high-frequency components
of the training set. 
\begin{center}
\begin{figure*}
\begin{centering}
\includegraphics[scale=0.45]{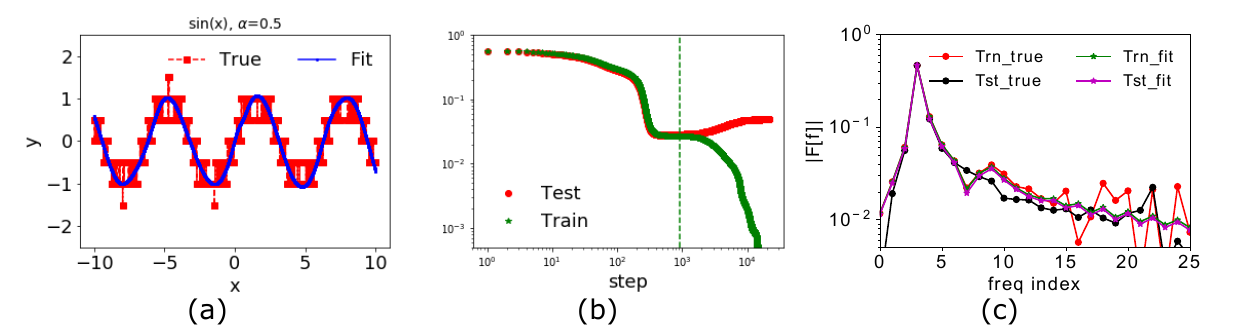}
\par\end{centering}
\caption{Effect of Early-stopping on contaminated data. The training set and
the test set consist of $300$ and $6000$ data points evenly sampled
in $[-10,10]$, respectively. (a) The sampled values of the test set
(red square dashed line) and DNN outputs (blue solid line) at the
turning step. (b) Loss functions for training set (green stars) and
test set (red dots) at different recording steps. The green dashed
line is drawn at the turning step, where the best generalization performance
is achieved. (c) $|\mathcal{F}[f]|$ for the training set (red) and
test set (black), and $|\mathcal{F}[T]|$ for the training set (green),
and test set (magenta) at the turning step. \label{fig:Generalization}}
 
\end{figure*}
\par\end{center}

\section{Conclusions and discussion}

In this work, we empirically discover an F-Principle underlying the
optimization process of DNNs. Specifically, for functions with dominant
low-frequency components, a DNN with common settings first capture
their low-frequency components while keeping its own high-frequency
ones small. In our experiments, this phenomenon can be widely observed
for DNNs of different width (tens to thousands in each layer), depth
(one to tens), training algorithms (GD, SGD, Adam), and activation
functions (tanh and ReLU). The F-Principle provides insights into
the good generalization performance of DNNs often observed in experiments.
In Appendix \ref{subsec:Compression-vs.-no}, we also discuss how
the F-Principle helps understand the training behavior of DNNs in
the information plane \cite{shwartz2017opening} . 

Note that initial parameters with large values could complicate the
phenomenon of the F-Principle. In previous experiments, the training
behavior of DNNs initialized by Gaussian distribution with mean $0$
and small standard deviation follows the F-Principle. However, with
large initialization, i.e., parameters initialized by a Gaussian distribution
of large standard deviation, it is difficult to observe a clear phenomenon
of the F-Principle. More importantly, these two initialization strategies
could result in very different generalization performances. When the
standard deviation for initialization is large \footnote{The bias terms are always initialized by standard deviation $0.1$.},
say, $10$ (see Fig.\ref{fig:Init}a), the initial DNN output fluctuates
strongly. In contrast, when the parameters of the DNN are initialized
with small values, say, Gaussian distribution with standard deviation
$0.1$, the initial DNN output is flat (see Fig.\ref{fig:Init}d).
For both initializations, DNNs can well fit the training data (see
Fig.\ref{fig:Init}b and e). However, for test data, the DNN with
small initialization generalizes well (Fig.\ref{fig:Init}f) whereas
the DNN with large initialization clearly overfits (Fig.\ref{fig:Init}c).
Intuitively, the above phenomenon can be understood as follows. Without
explicit constraints on the high-frequency components beyond the effective
frequency range of the training data, the DNN output after training
tends to inherit these high-frequency components from the initial
output. Therefore, with large initialization, the DNN output can easily
overfit the training data with fluctuating high-frequency components.
In practice, the parameters of DNNs are often randomly initialized
with standard deviations close to zero. As suggested by our analysis,
the small-initialization strategy may implicitly lead to a more efficient
and well-generalized optimization process of DNNs as characterized
by the F-Principle. Note that a quantitative study of how initialization
affects the generalization of DNN can be found in a subsequent work
\cite{zhang_type_2019}.
\begin{center}
\begin{figure}[h]
\begin{centering}
\includegraphics[scale=0.42]{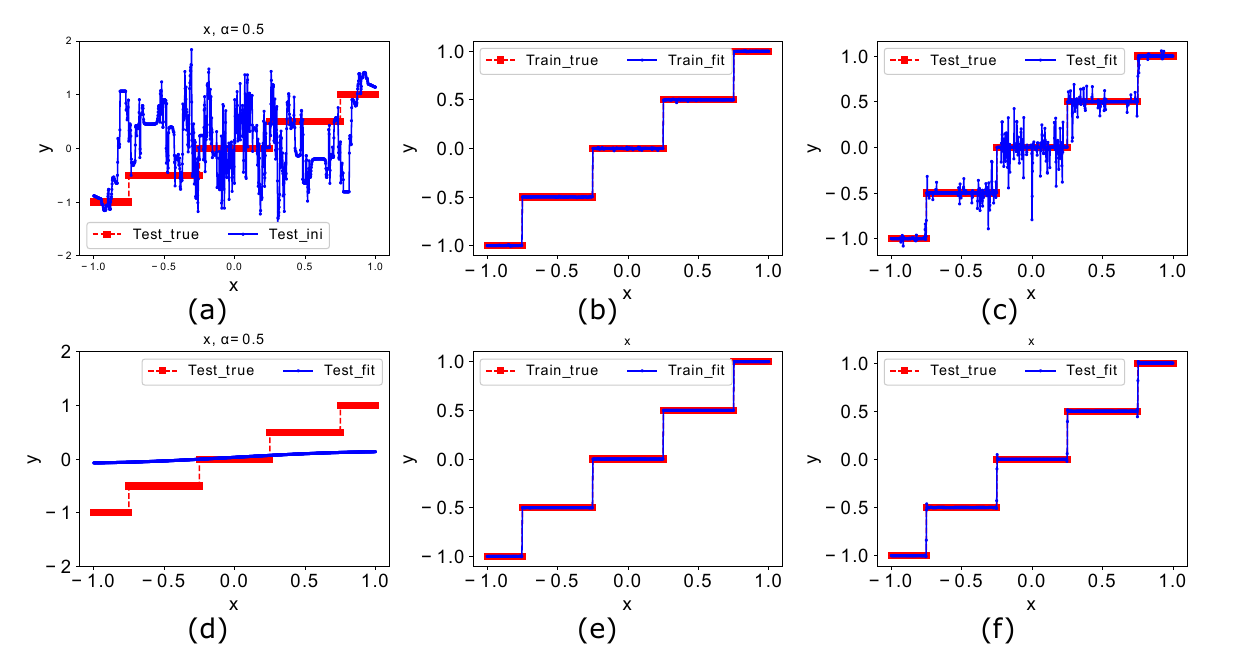}
\par\end{centering}
\caption{DNN outputs with different initializations for fitting function $f(x)$
of $f_{0}(x)=x$ with $\alpha=0.5$ in Eq. (\ref{eq:fittingFunc}).
The training data and the test data are evenly sampled in $[-1,1]$
with sample size $600$ and $1200$, respectively. The parameters
of DNNs are initialized by a Gaussian distribution with mean $0$
and standard deviation either $10$ (first row) or $0.1$ (second
row). (a, d): $f(x)$ (red dashed line) and initial DNN outputs (blue
solid line) for the test data. (b, e): $f(x)$ (red dashed line) and
DNN outputs (blue solid line) for the training data at the end of
training. (c, f): $f(x)$ (red dashed line) and DNN outputs (blue
solid line) for the test data at the end of training. \label{fig:Init}}
 
\end{figure}
\par\end{center}

\section*{Acknowledgments}

The authors want to thank David W. McLaughlin for helpful discussions
and thank Qiu Yang (NYU), Zheng Ma (Purdue University), and Tao Luo  (Purdue University), Shixiao Jiang (Penn State), Kai Chen (SJTU) for critically reading the manuscript. Part of this work was done when ZX, YZ, YX are postdocs at New York University Abu Dhabi and visiting members at Courant Institute supported by the NYU Abu Dhabi Institute
G1301. The authors declare no conflict of interest.

\subsection*{\bibliographystyle{splncs04}
}
\bibliography{DLRef}

\section{Appendix \label{subsec:Compression-vs.-no}}

Through the empirical exploration of the training behavior of DNNs
in the information plane, regarding information compression phase,
Schwartz-Ziv and Tishby (2017) \cite{shwartz2017opening} claimed
that (i) information compression is a general process; (ii) information
compression is induced by SGD. In this section, we demonstrate how
the F-Principle can be used to understand the compression phase.

\subsection{Computation of information\label{subsec:Mutual-information}}

For any random variables $U$ and $V$ with a joint distribution $P(u,v)$:
the entropy of $U$ is defined as $I(U)=-\sum_{u}P(u)\log P(u)$;
their mutual information is defined as $I(U,V)=\sum_{u,v}P(u,v)\log\frac{P(u,v)}{P(u)P(v)}$;
the conditional entropy of $U$ on $V$ is defined as 
\[
I(U|V)=\sum_{u,v}P(u,v)\log\frac{P(v)}{P(u,v)}=I(U)-I(U,V).
\]

By the construction of the DNN, its output $T$ is a deterministic
function of its input $X$, thus, $I(T|X)=0$ and $I(X,T)=I(T)$.
To compute entropy numerically, we evenly bin $X$, $Y$, $T$ to
$X_{b}$, $Y_{b}$, $T_{b}$ with bin size \emph{$b$} as follows.
For any value $v$, its binned value is define as $v_{b}={\rm Round}(v/b)\times b$.
In our work, $I(T)$ and $I(Y,T)$ are approximated by $I(T_{b})$
and $I(Y_{b},T_{b})$, respectively, with $b=0.05$. Note that, after
binning, one value of $X_{b}$ may map to multiple values of $T_{b}$.
Thus, $I(T_{b}|X_{b})\neq0$ and $I(X_{b},T_{b})\neq I(T_{b})$. The
difference vanishes as bin size shrinks. Therefore, with a small bin
size, $I(T_{b})$ is a good approximation of $I(X,T)$. In experiments,
we also find that $I(X_{b},T_{b})$ and $I(T_{b})$ behave almost
the same in the information plane for the default value $b=0.05$.

\subsection{Compression vs. no compression in the information plane}

We demonstrate how compression can appear or disappear by tuning the
parameter $\alpha$ in Eq. (\ref{eq:fittingFunc}) with $f_{0}(x)=x$
for $x\in[-1,1]$ using full batch gradient descent (GD) without stochasticity.
In our simulations, the DNN well fits $f(x)$ for both $\alpha$ equal
to $0$ and $0.5$ after training (see Fig.\ref{fig:ERJnum}a and
c). In the information plane, there is no compression phase for $I(T)$
for $\alpha=0$ (see Fig.\ref{fig:ERJnum}b). By increasing $\alpha$
in Eq. (\ref{eq:fittingFunc}) we can observe that: i) the fitted
function is discretized with only few possible outputs (see Fig.\ref{fig:ERJnum}c);
ii) the compression of $I(T)$ appears (see Fig.\ref{fig:ERJnum}d).
For $\alpha>0$, behaviors of information plane are similar to previous
results \cite{shwartz2017opening}. To understand why compression
happens for $\alpha>0$, we next focus on the training courses for
different $\alpha$ in the frequency domain. 

A key feature of the class of functions described by Eq. (\ref{eq:fittingFunc})
is that the dominant low-frequency components for $f(x)$ with different
$\alpha$ are the same. By the F-Principle, the DNN first captures
those dominant low-frequency components, thus, the training courses
for different $\alpha$ at the beginning are similar, i.e., i) the
DNN output is close to $f_{0}(x)$ at certain training epochs (blue
lines in Fig.\ref{fig:ERJnum}a and c); ii) $I(T)$ in the information
plane increases rapidly until it reaches a value close to the entropy
of $f_{0}(x)$ , i.e., $I(f_{0}(x))$ (see Fig.\ref{fig:ERJnum}b
and d). For $\alpha=0$, the target function is $f_{0}(x)$, therefore,
$I(T)$ will be closer and closer to $I(f_{0}(x))$ during the training.
For $\alpha>0$, the entropy of the target function, $I(f(x))$, is
much less than $I(f_{0}(x))$. In the latter stage of capturing high-frequency
components, the DNN output $T$ would converge to the discretized
function $f(x)$. Therefore, $I(T)$ would decrease from $I(f_{0}(x))$
to $I(f(x))$. 

This analysis is also applicable to other functions. As the discretization
is in general inevitable for classification problems with discrete
labels, we can often observe information compression in practice as
described in the previous study \cite{shwartz2017opening}. 
\begin{center}
\begin{figure}[h]
\begin{centering}
\includegraphics[scale=0.8]{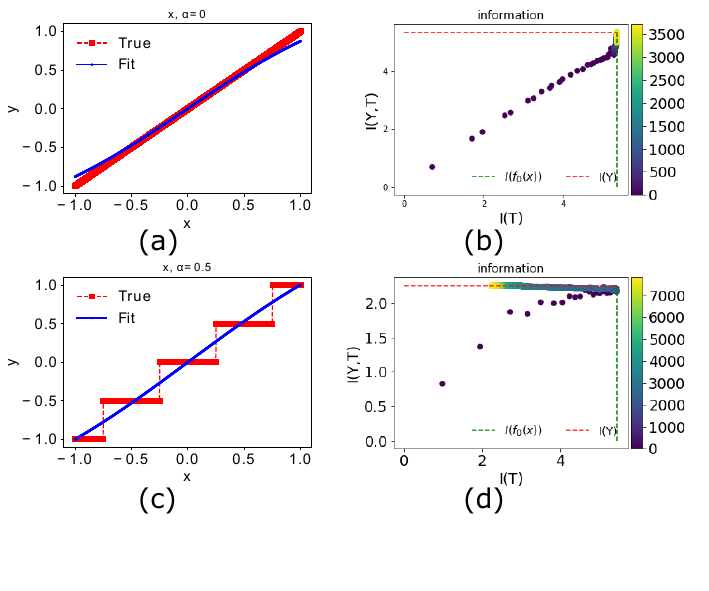}
\par\end{centering}
\caption{Analysis of compression phase in the information plane. $\alpha$
is $0$ for (a, b) and $0.5$ for (c, d). (a, c) $f(x$) (red square)
with $f_{0}(x)=x$ in Eq. (\ref{eq:fittingFunc}) and the DNN output
(blue solid line) at a certain training step. (b, d) Trajectories
of the training process of the DNN in the information plane. Color
of each dot indicates its recording step. The green dashed vertical
line and the red dashed horizontal line indicate constant values of
$I(f_{0}(x))$ and $I(Y)$, respectively. \label{fig:ERJnum}}
 
\end{figure}
\par\end{center}
\end{document}